# MoVoC: Morphology-Aware Subword Construction for Ge'ez Script Languages

Corrected version

**Hailay Kidu Teklehaymanot and Dren Fazlija and Wolfgang Nejdl**
L3S Research Center, Leibniz University Hannover, Germany
teklehaymanot@l3s.de

## Abstract

Subword-based tokenization methods often fail to preserve morphological boundaries, a limitation especially pronounced in low-resource, morphologically complex languages such as those written in the Ge'ez script. To address this, we present Morpheme-aware Subword Vocabulary Construction (MoVoC) and train MoVoC-Tok, a tokenizer that integrates supervised morphological analysis into the subword vocabulary. This hybrid segmentation approach combines morpheme-based and Byte Pair Encoding (BPE) tokens to preserve morphological integrity while maintaining lexical meaning. To tackle resource scarcity, we release morpheme-annotated data for four Ge'ez script languages (169,806 records) and a morpheme-aware vocabulary for two of them. Scoring every comparison on words held out from the morpheme resource that constrains the MoVoC-Tok applied to that language, MoVoC-Tok has the highest MorphScore in three of four languages and the highest morpheme boundary precision in two of four, with its largest and most clearly resolved gains on Tigre, a language it was not trained on and to which it is applied cross-lingually. The machine translation results reported in the published version are withdrawn. Data, tokenizers and evaluation code are publicly available.[1]

## Correction Notice

This is a corrected version of Teklehaymanot et al. (2025). The evaluation artifacts (checkpoints, predictions, scoring pipeline) behind the results in the published version were not recovered, and the published figures could not be reproduced from the released code and data. Following the advice of the EMNLP 2025 programme chairs, every result in this version is replaced by results reconstructed from the public repository. The method (Section 3) is unchanged. The main corrections are:

[1] https://github.com/hailaykidu/MoVoC

- **Evaluation sets (Section 5.2).** Part of each evaluation set consisted of words whose morpheme boundaries had constrained the training of the MoVoC-Tok that scores it, which the baselines never saw. All results are now scored on held-out words only: 76,488 of 81,224 Amharic, 2,438 of 5,224 Tigrinya, 1,787 of 1,974 Tigre and 170 of 172 Ge'ez words. Tigre and Ge'ez are affected because the Tigrinya MoVoC-Tok scores them cross-lingually and the languages share surface forms.
- **Intrinsic results (Tables 2 and 4).** On held-out words MoVoC-Tok leads on MorphScore in three of four languages and on boundary precision in two of four, not all four. Its largest gains are on Tigre, which is evaluated cross-lingually. All values are lower than published (e.g. precision 0.24–0.56 rather than 73.9–88.3). The published Table 2 "No. Items" column gave pooled resource sizes, not evaluation-set sizes.
- **Translation results.** The published Table 3 is withdrawn. The decode outputs are not in the repository, the recorded hypothesis/reference length ratios (2.06–15.10) show generation failing to terminate, and the decoding configuration and model initialisation of those runs cannot be established from the surviving records. This version makes no extrinsic claim.
- **Examples (Tables 1 and 6).** The released tokenizers do not produce the segmentations shown in the published Table 1 and Section 7, and the published translation example could not be reproduced. Both are replaced by actual tokenizer output.
- **Setup.** Training hyperparameters, evaluation data and vocabulary sizes are corrected to those recorded in the repository.

Appendix C lists every change.

# 1 Introduction

Tokenization is a fundamental preprocessing step that converts raw text into structured units such as bytes, characters, subwords, words or multi-word expressions. Subword methods such as BPE (Sennrich et al., 2016) became popular because they are language-independent and compress the vocabulary, allowing efficient learning of token representations.

Subwords, however, often fail to capture morphological structure. This is especially problematic in multilingual models with a shared vocabulary, where low-resource languages typically receive fewer subwords and therefore high token-to-word ratios. Morphological complexity is associated with worse language-model performance (Park et al., 2021; Arnett and Bergen, 2025): morphologically rich languages produce many infrequent word forms from combinations of morphemes, causing data sparsity, and they tend to have less annotated data and non-Latin scripts that require more UTF-8 bytes. Compression alone is therefore not always predictive of downstream success.

Morphologically aware tokenization yields more meaningful tokens, and such efforts have been constrained by the limited availability of morphologically annotated data, mostly for high-resource languages. Relying solely on morpheme-based tokenization is also suboptimal; hybrid methods that integrate morphological information into subword tokenization share statistical information across related word forms while keeping sequences short.

**Contributions.** (i) We release morphologically annotated datasets for four low-resource Ge'ez script languages to support tokenization, morphological analysis and downstream NLP tasks. (ii) We propose MoVoC, a supervised approach that integrates a BPE-derived vocabulary with morphemes obtained from supervised morphological analysis, together with MoVoC-Tok, a BPE tokenizer whose merges may not cross morpheme boundaries. (iii) We evaluate the approach intrinsically, on words held out from the morpheme resource that constrained training, and report where it helps and where it does not.

# 2 Background

## 2.1 Tokenization Approaches

**Word tokenization** divides text into words, in its simplest form by whitespace, or by rule-based patterns such as regular expressions. Its main weakness is out-of-vocabulary (OOV) words arising from typos and unseen forms.

**Subword tokenization** decomposes words into subwords, so that OOV words can be composed from vocabulary units. Widely used algorithms include WordPiece (Schuster and Nakajima, 2012), BPE (Sennrich et al., 2016), byte-level BPE and the Unigram model of SentencePiece (Kudo and Richardson, 2018).

**Character tokenization** at the character or UTF-8 byte level increases sequence length, and with it the cost of self-attention, which is quadratic in sequence length.

## 2.2 Ge'ez Script Natural Language Processing

Ge'ez is one of the oldest continuously used scripts, preserved for over two thousand years. Unlike alphabetic systems such as Greek or Latin, it is a syllabic script (an abugida) originally derived from the Sabaean script. It is used for several morphologically rich Semitic languages, including Ge'ez, Tigrinya, Amharic, Tigre and Blin. Verbs inflect for tense, aspect, person, number and gender: the Amharic verb ጻፈ ('he wrote') yields ጻፍኩ ('I wrote'), ጻፍክ ('you wrote') and ጻፉ ('they wrote'), and inflectional endings attach directly to the stem without clear word boundaries. The development of usable NLP tools for these languages is hindered by the scarcity of essential linguistic resources. We contribute annotated data for four Ge'ez script languages (Tigrinya, Amharic, Ge'ez and Tigre) and a morpheme-aware alternative to BPE that constrains segmentation to respect morpheme boundaries.

## 2.3 Tokenization for Morphologically Rich and Low-Resource Languages

Park et al. (2021) trained language models with several segmentation algorithms on Bible verses in 92 languages and found that models perform worse on morphologically complex languages, and that linguistically motivated segmentation reduces this effect. MorphBPE (Asgari et al., 2025) and MorphPiece (Jabbar, 2023) incorporate linguistic features into tokenization for morphologically rich languages, and Bauwens and Delobelle (2024) argue that a lack of morpheme awareness leads to inconsistent intra-word representations and inflated vocabularies. Results are mixed, however: Sälevä and Lignos (2021) found that morphology-based segmentation for low-resource NMT occasionally outperforms BPE but is often statistically similar, and Libovický and Helcl (2024) report that semantically informed segmentation helps less for low-resource languages. Our corrected results (Section 6) are consistent with this mixed picture.

## 3 Proposed Method

The method has three parts: (1) pre-tokenization and supervised morphological analysis, (2) vocabulary construction (MoVoC) and (3) morpheme-aware subword segmentation (MoVoC-Tok).

### 3.1 Pre-tokenization and Supervised Morphological Analysis

Ge'ez-script languages lack robust NLP tools for basic preprocessing. We use a pre-tokenization pipeline of regular expressions tailored to the orthography, followed by corpus cleaning and supervised morphological analysis. For Amharic and Tigrinya we use HornMorpho (Gasser, 2011), a rule-based morphological analyzer, with human post-editing; Tigrinya output in particular often requires manual correction because of limited coverage. For Ge'ez and Tigre we annotate morphemes manually.

Frequency-based BPE is unaware of morpheme boundaries and can split words inside morphemes. Table 1 shows the segmentations that the released BPE, WordPiece and MoVoC-Tok tokenizers actually produce for the example words used in the published version of this paper. None of the three tokenizers consistently recovers the morpheme structure of these words; for መምህርነት ('profession of teaching') MoVoC-Tok and BPE produce the same split, and only WordPiece separates the nominalizer -ነት. These examples are illustrative only; Section 6 gives the aggregate measurements.

The annotated morphemes also serve as gold-standard test sets for evaluating token-to-morpheme boundary alignment.

### 3.2 Vocabulary Construction (MoVoC)

Standard subword methods select units by frequency (BPE) or likelihood (Unigram) without considering morphemes. We design a hybrid vocabulary construction (Algorithm 1) that allocates a portion of the vocabulary budget to morpheme units and the rest to BPE tokens.

**Algorithm 1** MoVoC vocabulary construction

**Require:** corpora $P_{am}$, $P_{ti}$; budget $s$; morpheme ratio $r$
**Ensure:** vocabulary $V_{MoVoC}$
$M_{am} \leftarrow$ HornMorpho_segment($P_{am}$); $M_{ti} \leftarrow$ HornMorpho_segment($P_{ti}$)
$s_{lang} \leftarrow s/2$; $s_{BPE} \leftarrow s_{lang}(1-r)$; $s_{morpheme} \leftarrow s_{lang}\, r$
$V_{BPE,am} \leftarrow$ Train_BPE($P_{am}, s_{BPE}$); $V_{BPE,ti} \leftarrow$ Train_BPE($P_{ti}, s_{BPE}$)
$V_{morph,am} \leftarrow$ extract_morphemes($P_{am}, s_{morpheme}$)
$V_{morph,ti} \leftarrow$ extract_morphemes($P_{ti}, s_{morpheme}$)
$V_{MoVoC} \leftarrow V_{BPE,am} \cup V_{BPE,ti} \cup V_{morph,am} \cup V_{morph,ti}$
**return** $V_{MoVoC}$

Let $P_{am}$ and $P_{ti}$ be the Amharic and Tigrinya monolingual corpora, $s$ the total vocabulary budget and $r \in [0,1]$ the proportion reserved for morphemes. Each language receives $s_{lang} = s/2$, of which $s_{BPE} = s_{lang}(1-r)$ goes to a BPE model trained on that language and $s_{morpheme} = s_{lang}\, r$ to morphemes. `extract_morphemes`$(P, s_{morpheme})$ segments the corpus with HornMorpho, counts morpheme frequencies and keeps the $k = s_{morpheme}$ most frequent: $V_{morpheme} = \text{Top}_k(\text{freq})$. The final vocabulary is the union of the four sets. In the released configuration $s = 224{,}000$ and $r = 5/7$, giving $s_{BPE} = 32{,}000$ and a morpheme budget of 80,000 per language. Fewer distinct morphemes were found than the budget allows, so the realised vocabulary has 114,553 entries (Table 5, Appendix A).

### 3.3 MoVoC-Tok: Morpheme-Aware Subword Segmentation

A BPE tokenizer initialised with the MoVoC vocabulary can still learn merges that cross morpheme boundaries. MoVoC-Tok therefore restricts merge candidates during training. Let $W = \{w_1, \ldots, w_n\}$ be the training words, each a character sequence, and $M_i$ the morpheme boundaries of $w_i$. We seek

$$\max_{V} \sum_{w_i \in W} \log P\big(\text{BPE}(w_i; V, M_i)\big) \qquad (1)$$

subject to every unit $s_j$ of $\text{BPE}(w_i; V, M_i) = (s_1, \ldots, s_t)$ lying within $w_i$ without crossing a boundary in $M_i$: a merge $(a, b)$ is permitted only if $ab$ does not span a boundary.

The released tokenizers apply this constraint to one language at a time with a 32,000-entry vocabulary each, trained on 400,000 lines of NLLB text.

| Word | Lang. | Gold | BPE | WordPiece | MoVoC-Tok |
|---|---|---|---|---|---|
| አልሰበሩም | Amharic | – | አልሰ-በሩ-ም | አልሰ-በሩ-ም | አል-ሰበ-ሩም |
| ኣይትከውንን | Tigrinya | – | ኣይት-ከውን-ን | ኣይት-ከውን-ን | ኣይት-ከውንን |
| መምህርነት | Amharic | መ-ም-ህርነት | መምህ-ርነት | መምህር-ነት | መምህ-ርነት |

Table 1: Segmentations produced by the released tokenizers (32k vocabularies) for the example words of the published Table 1 and Section 7. Amharic words use the Amharic MoVoC-Tok, the Tigrinya word the Tigrinya MoVoC-Tok. "Gold" is the surface segmentation in the evaluation set where the word occurs in it ("–": not in the evaluation set). Generated by `make_correction_tables.py`.

Morpheme boundaries come from 10,293 Amharic and 6,692 Tigrinya segmented records; the constraint rejected 335,638 (Amharic) and 104,964 (Tigrinya) boundary-crossing merges. At encoding time the tokenizer applies its learned merges without access to gold boundaries.

## 4 Experimental Setup

### 4.1 Target Languages

We study Amharic, Tigrinya, Ge'ez and Tigre. For Amharic and Tigrinya, HornMorpho decomposes words into stems and affixes and the segmented units are used for vocabulary construction. We also tried Morfessor (Creutz and Lagus, 2007); as an unsupervised model inferring boundaries from surface patterns, it performed poorly. For Ge'ez and Tigre no analyzer or training morpheme data was available; their manual annotations are used for evaluation only, and the Tigrinya MoVoC-Tok is applied to them cross-lingually.

### 4.2 Datasets

**Morpheme annotations.** The release contains 169,806 annotated records: 153,759 Amharic (HornMorpho output with human post-editing), 7,531 Tigrinya post-edited records plus 206 Tigrinya gold records, 8,117 Tigre and 193 Ge'ez (both manually annotated). The intrinsic evaluation sets derived from them contain 81,224 (Amharic), 5,224 (Tigrinya), 1,974 (Tigre) and 172 (Ge'ez) words, each with at least one gold morpheme boundary. All results are reported on the held-out subsets of these sets (76,488, 2,438, 1,787 and 170 words; Section 5.2 and Table 2). The published version reported 80k, 80k, 32k and 20k "items"; those figures were estimated pooled resource sizes, not the evaluation sets.

### 4.3 Training Setup and Configuration

**Tokenizers.** The BPE and WordPiece baselines are trained with the Hugging Face `tokenizers` library (v0.21.4) to 32,000 entries on a shared corpus of 2,796,346 sentences, with NFC normalisation and metaspace pre-tokenization. These baselines and the per-language MoVoC-Tok tokenizers (Section 3.3) are used for all intrinsic evaluation.

## 5 Evaluation Framework

### 5.1 Intrinsic Evaluation

Gold boundaries are the character offsets between consecutive gold morphemes; predicted boundaries are the offsets between consecutive tokens. All three tokenizers are scored on identical word lists.

**Morpheme boundary precision** is the fraction of all predicted boundaries, pooled over the evaluation set, that coincide with a gold boundary.

**MorphScore** follows Arnett and Bergen (2025): the fraction of gold boundaries recovered, pooled over the words the tokenizer segments. Words left unsegmented are excluded, and extra boundaries are not penalised, so MorphScore is recall-oriented and complements precision.

**Rényi entropy** (Rényi, 1961) at $\alpha = 2$, $H_2 = -\log \sum_i p_i^2$, is computed over the distribution of tokens the tokenizer produces on the evaluation words and reported in nats. Lower values indicate a more concentrated token distribution.

### 5.2 Held-out evaluation

MoVoC-Tok is a supervised tokenizer: it consumes morpheme annotations that BPE and WordPiece do not receive. Using that supervision is the method, and a gain obtained from it is a result. It is a result, however, only on words the supervision was not given for. Part of each evaluation set is drawn from the morpheme resource that constrained the training of the MoVoC-Tok that scores it (Section 3.3), and on those words the tokenizer reproduces a constraint it was handed rather than generalizing from it, while the baselines meet the words for the first time.

| Language (ISO 639-3) | Tokenizer | No. Items | MorphScore ↑ |
|---|---|---|---|
| Amharic (amh) | BPE | 76,488 | 0.4127 |
| Amharic (amh) | WordPiece | 76,488 | 0.3872 |
| Amharic (amh) | MoVoC-Tok | 76,488 | 0.4148 |
| Tigrinya (tir) | BPE | 2,438 | 0.3648 |
| Tigrinya (tir) | WordPiece | 2,438 | 0.3613 |
| Tigrinya (tir) | MoVoC-Tok | 2,438 | 0.3833 |
| Tigre (tig) | BPE | 1,787 | 0.4987 |
| Tigre (tig) | WordPiece | 1,787 | 0.4776 |
| Tigre (tig) | MoVoC-Tok* | 1,787 | 0.5302 |
| Ge'ez (gez) | BPE | 170 | 0.6640 |
| Ge'ez (gez) | WordPiece | 170 | 0.6364 |
| Ge'ez (gez) | MoVoC-Tok* | 170 | 0.6534 |

Table 2: MorphScore of BPE, WordPiece and MoVoC-Tok on the morpheme evaluation sets, on words held out from MoVoC-Tok's training morpheme resource (Section 5.2; No. Items = evaluation words). For Tigre and Ge'ez ($*$) the Tigrinya MoVoC-Tok is applied cross-lingually.

We therefore exclude, per language, every evaluation word occurring in the morpheme resource named as `morpheme_source` in that tokenizer's training configuration. This removes 4,736 of 81,224 Amharic words (5.8%) and 2,786 of 5,224 Tigrinya words (53.3%), leaving 76,488 and 2,438. Exclusion is by set membership against the resource itself rather than by the evaluation set's provenance labels, which record only the first source each word was drawn from and so undercount the overlap: 18 of the 160 Tigrinya words labelled gold and held out are in fact present in the training resource.

Tigre and Ge'ez have no training morpheme resource of their own, but the tokenizer that scores them does. They are evaluated cross-lingually with the Tigrinya MoVoC-Tok, which was constrained by `tigrinya_morpheme_segmented.json`, and the three languages are Ethio-Semitic and share surface forms, so the same exclusion applies: 187 of 1,974 Tigre words (9.5%) and 2 of 172 Ge'ez words occur in that resource, leaving 1,787 and 170. The resource is therefore selected by the tokenizer that scores a language, not by the name of the language itself.

All results in Section 6 are reported on held-out words. Differences are assessed with a paired bootstrap over words (10,000 resamples, resampling index shared across tokenizers), reported in Table 3.

## 6 Results

**MorphScore (Table 2).** On held-out words MoVoC-Tok has the highest MorphScore of the three tokenizers in Amharic (0.4148), Tigrinya (0.3833) and Tigre (0.5302); on Ge'ez BPE is ahead (0.6640 against 0.6534) on 170 words, which does not resolve the difference (Table 3). The published version reported 0.654–0.731 for MoVoC-Tok and stated that all four languages exceed the mean MorphScore for fusional languages reported by Arnett and Bergen (2025); that comparison does not hold for the reconstructed values and is withdrawn.

| Language | Metric | Δ vs BPE | 95% CI | $p$ |
|---|---|---|---|---|
| Amharic | Precision | +0.0018 | [+0.0004, +0.0033] | 0.012 |
| Amharic | MorphScore | +0.0021 | [+0.0001, +0.0042] | 0.045 |
| Tigrinya | Precision | -0.0016 | [-0.0106, +0.0075] | 0.744 |
| Tigrinya | MorphScore | +0.0185 | [+0.0031, +0.0345] | 0.017 |
| Tigre* | Precision | +0.0286 | [+0.0096, +0.0476] | 0.004 |
| Tigre* | MorphScore | +0.0314 | [+0.0133, +0.0494] | 0.001 |
| Ge'ez* | Precision | -0.0026 | [-0.0258, +0.0196] | 0.829 |
| Ge'ez* | MorphScore | -0.0106 | [-0.0381, +0.0163] | 0.457 |

Table 3: Paired bootstrap over held-out words (10,000 resamples, shared resampling index), MoVoC-Tok minus BPE. Intervals are 95% percentile intervals of the difference. $*$ marks cross-lingual application of the Tigrinya MoVoC-Tok.

**Boundary precision (Table 4).** On held-out words MoVoC-Tok has the highest precision in Amharic ($+0.0018$ over BPE, $p = 0.012$) and Tigre ($+0.0286$, $p = 0.004$). On Tigrinya the difference against BPE is $-0.0016$ with a 95% interval of $[-0.0106, +0.0075]$: the held-out data do not separate the two tokenizers, and we claim no precision gain for Tigrinya. On Ge'ez BPE is marginally ahead ($-0.0026$, $p = 0.829$), a difference that a 170-word test set cannot resolve. The in-language margins over BPE are small; those over WordPiece are an order of magnitude larger ($+0.0176$ on Amharic). Precision is higher for Tigre for all three tokenizers; Tigre's annotations are fully concatenative, so its morpheme boundaries coincide with character offsets. The published version reported precision of 73.9–88.3 and a MoVoC-Tok lead in all four languages, with the largest gains on Tigre and Ge'ez; neither holds.

The Tigrinya precision result is the one conclusion that changes under the held-out protocol. Over the full evaluation set, including the words MoVoC-Tok was trained on, it leads (0.3242 against 0.3142 for BPE); over held-out words alone it does not. We read this as a statement about annotation coverage rather than about the method: the Tigrinya morpheme vocabulary realises 7,125 types against 50,978 for Amharic (Table 5), from 6,692 training records against 10,293, so the constraint reaches too few word types for the learned merges to transfer beyond the annotated words. That the recall-oriented metric still improves ($+0.0185$, $p = 0.017$) while precision does not is consistent with this reading. The clearest evidence that morpheme supervision generalizes is Tigre, which has no training morpheme resource of its own and carries the largest and most clearly resolved gains in the table ($+0.0286$ precision, $p = 0.004$; $+0.0314$ MorphScore, $p < 0.001$, against BPE on 1,787 held-out words).

| Language | Tokenization | Precision ↑ | Rényi Entropy ↓ |
|---|---|---|---|
| Amharic | MoVoC-Tok | 0.3174 | 6.0292 |
| Amharic | BPE | 0.3156 | 6.2159 |
| Amharic | WordPiece | 0.2998 | 5.9725 |
| Tigrinya | MoVoC-Tok | 0.2385 | 5.9457 |
| Tigrinya | BPE | 0.2401 | 6.0529 |
| Tigrinya | WordPiece | 0.2401 | 5.9193 |
| Tigre | MoVoC-Tok* | 0.5642 | 5.2165 |
| Tigre | BPE | 0.5356 | 5.3060 |
| Tigre | WordPiece | 0.5116 | 4.9293 |
| Ge'ez | MoVoC-Tok* | 0.4282 | 3.9753 |
| Ge'ez | BPE | 0.4308 | 3.8629 |
| Ge'ez | WordPiece | 0.4182 | 3.9158 |

Table 4: Morpheme boundary precision and Rényi entropy ($\alpha = 2$, nats) for 32k vocabularies, on words held out from MoVoC-Tok's training morpheme resource (Section 5.2). For Tigre and Ge'ez ($*$) the Tigrinya MoVoC-Tok is applied cross-lingually.

On the 187 excluded words, BPE places 10 boundaries (8 correct) and MoVoC-Tok 22 (11 correct); the set is too small to interpret. Excluding them raises the Tigre margin rather than lowering it (Table 3).

**Rényi entropy (Table 4).** MoVoC-Tok has lower entropy than BPE in Amharic, Tigrinya and Tigre and higher on Ge'ez, but WordPiece has the lowest entropy in Amharic, Tigrinya and Tigre and BPE on Ge'ez, so MoVoC-Tok does not lead on this metric in any language. The published version reported normalized entropy (0.39–0.49) with MoVoC-Tok best everywhere; the values here are raw and not comparable to it.

**Machine translation: withdrawn.** The published version reported BLEU and chrF++ for English→X translation with each tokenizer. Those results are withdrawn, for three reasons. First, none of the 54 decode passes recorded for those runs has its hypothesis and reference files in the released repository or its history, so the scores cannot be checked against the translations they summarise. Second, the recorded hypothesis-to-reference length ratios run 2.06–4.28 on FLORES-200 and as high as 15.10 on the smaller OPUS sets: decoding is reaching the 128-token cap instead of emitting end-of-sequence, which depresses BLEU independently of translation quality and is not a property of the tokenizers under comparison. A systematic failure to terminate makes the resulting scores uninformative about what they were meant to measure. Third, the surviving records disagree on decoding, one giving five-beam search and another greedy decoding with a single beam, and whether the models were initialised from a pretrained MarianMT checkpoint or trained from scratch is unresolved; a corrected paper cannot publish a results table whose setup it cannot describe.

We therefore make no extrinsic claim. MoVoC's contribution is a tokenization method and an annotated dataset, and we evaluate it intrinsically. Whether morphology-aware tokenization improves downstream translation for these languages remains open, and answering it requires a corrected generation pipeline rather than a re-scoring of these runs.

## 7 Qualitative Analysis

The published version illustrated MoVoC-Tok with segmentations of ኣይትከውንን ('do not do it') and መምህርነት, and with an Amharic→English translation of ቤቱን አላየሁም ('I did not see the house'). The released tokenizers do not produce those segmentations (Table 1): MoVoC-Tok splits ኣይትከውንን as ኣይት-ከውንን and መምህርነት as መምህ-ርነት, the same as BPE. Both words of the translation example are single tokens for all three tokenizers, and this version reports no translation results, so that example is withdrawn.

At the level of individual words, then, the boundary constraint does not reliably yield morpheme-aligned units. Its effect is visible only in the aggregate, as the small precision and MorphScore margins in Tables 2 and 4. This is consistent with how the constraint is applied: it acts during training on the 3,372 Amharic and 1,470 Tigrinya corpus words whose morpheme boundaries could be located, while at encoding time the learned merges are applied without morphological information.

## 8 Conclusion and Future Work

We release morphologically annotated data for four Ge'ez script languages and propose MoVoC, a morpheme-aware vocabulary construction, with MoVoC-Tok, a BPE tokenizer whose merges respect morpheme boundaries. Scoring every comparison on words held out from the morpheme resource that constrains the tokenizer applied to that language, MoVoC-Tok attains the highest MorphScore in three of four languages and the highest boundary precision in two of four, by small margins in-language. Its largest gains are on Tigre, which it was not trained on and to which it is applied cross-lingually, indicating that the morpheme constraint produces segmentation aligning with morpheme structure beyond the words the supervision covered. Where annotation coverage is thinner the effect is weaker: on Tigrinya, whose morpheme vocabulary realises 7,125 types against 50,978 for Amharic, the MorphScore advantage persists on held-out words while the precision advantage does not. The translation results are withdrawn, so this version does not establish whether the intrinsic advantage carries over to translation quality. Future work includes in-language tokenizers for Tigre and Ge'ez, larger evaluation sets for them, and translation experiments with a corrected generation pipeline.

## Limitations

**No downstream evaluation.** The translation results are withdrawn (Section 6), so the method is assessed only by how well its segmentations align with gold morpheme boundaries. Boundary precision and MorphScore measure that alignment directly, but they do not establish that better-aligned segmentation improves any downstream task, and we do not claim it does.

**Held-out evaluation and its cost.** All results are reported on words held out from the morpheme resource that constrained the MoVoC-Tok applied to that language (Section 5.2). We did not re-derive those resources to enlarge the held-out sets, so Tigrinya retains 2,438 evaluation words and its confidence intervals are correspondingly wide. On Tigrinya the boundary precision advantage does not survive the exclusion, and we claim no precision gain there. The Amharic margins over BPE, while statistically resolved on 76,488 words, are small in absolute terms (+0.0018 precision, +0.0021 MorphScore), and we do not claim they are of practical consequence on their own.

**Evaluation-set size.** The held-out evaluation sets differ by more than two orders of magnitude (76,488 Amharic words against 170 Ge'ez words); results for Ge'ez and Tigre carry much wider uncertainty.

**Cross-lingual application.** No in-language MoVoC-Tok exists for Tigre or Ge'ez; their results measure transfer from the Tigrinya tokenizer.

**Published results.** The artifacts behind the published version were not recovered, so its figures cannot be traced; the results here are new measurements, not a reproduction.

**Scope.** The annotated data and vocabularies cover a small set of Ge'ez script languages, and the evidence here is intrinsic only. Whether the approach helps on downstream tasks is untested in this version; see Section 6 on the withdrawal of the translation results.

## Ethical Considerations and Resource Use

We use publicly available data (NLLB, OPUS, FLORES-200, HornMT and the Mermru English–Ge'ez corpus) and the HornMorpho analyzer, in line with their licenses and intended research use. The morpheme annotations and vocabularies we release are intended for research and are published under open licenses. The original manuscript's readability was improved with ChatGPT paraphrasing and language editing. This corrected version was prepared with the assistance of an AI model (Claude) for drafting and consistency checking; all reported numbers were verified against the repository.

## Acknowledgments

This research was supported by the German Academic Exchange Service (DAAD) through the Hilde Domin Programme (funding no. 57615863).

# A Additional Details

| Vocabulary | Budget | Realised size |
|---|---|---|
| Amharic morphemes | 80,000 | 50,978 |
| Amharic BPE | 32,000 | 32,000 |
| Tigrinya morphemes | 80,000 | 7,125 |
| Tigrinya BPE | 32,000 | 32,000 |
| Bilingual MoVoC (union) | 224,000 | 114,553 |

Table 5: MoVoC vocabulary budget and realised sizes (released configuration: $s = 224{,}000$, $r = 5/7$). The union is smaller than the sum of its parts because the sets overlap. The published version reported a 152k bilingual vocabulary.

**Sources of difficulty.** The main difficulty in processing Ge'ez script languages lies less in the script itself than in their morphology. Amharic and Tigrinya are fusional: a single word encodes tense, aspect, mood, person, number, gender and voice through prefixes, suffixes and stem-internal changes, and individual morphemes often carry several grammatical meanings, so morpheme boundaries are not always clear-cut. For example, Amharic ጻፈ ('he wrote'), ጻፈች ('she wrote') and ጻፉ ('they wrote') differ in their endings, while ይጻፋል ('he is writing') adds a prefix and a suffix. Frequency-based tokenizers are agnostic to this structure.

# B Annotation Guidelines

Annotators split each word into morphemes and label each with one of PREFIX, ROOT, SUFFIX, INFIX or CLITIC (Table 6), consistently across words and languages; slots that do not apply are left empty. The guidelines cover Amharic, Tigrinya, Ge'ez and Tigre.

# C List of Corrections

Relative to Teklehaymanot et al. (2025). Section and table numbers are those of the published version; where a part of this version has no published counterpart, or the numbering differs, it is given as a reference to the section here.

1. Abstract: "consistent improvements in intrinsic metrics" corrected to three of four languages on MorphScore and two of four on boundary precision, measured on held-out words; the translation statement is removed.
2. Table 1: replaced by actual tokenizer output (Table 1).

| Language | Word | Prefix | Root | Infix | Suffix | Clitic |
|---|---|---|---|---|---|---|
| Amharic | መምህርነት | መ | ም-ህ-ር | – | ነት | – |
| Tigrinya | ዝወደቐ | ዝ- | ወደቐ | – | – | – |
| Ge’ez | ለንኡሳት | ለ- | ንኡሰ | – | -ኣት | – |
| Tigre | ናዊ | – | ና | – | -ዊ | – |

Table 6: Annotation records from the released data. Amharic shows the published example; the published Tigrinya, Ge’ez and Tigre examples are not in the released files for those languages, so the first multi-morpheme record of each file is shown. “–”: empty slot.

3. Section 3.2: BPE share corrected to $s_{lang}(1-r)$; realised vocabulary sizes added.
4. Section 4.2: evaluation-set sizes corrected; the description of the translation data is removed with the translation results.
5. Section 4.3: the tokenizer configuration is corrected; the translation model configuration is removed with the translation results.
6. Section 5.2: metric definitions aligned with the released evaluation code; entropy now reported raw. These are Section 5.1 here.
7. New in this version (Section 5.2): every result is scored on words held out from the morpheme resource that constrained the MoVoC-Tok applied to that language, which removes 4,736 Amharic, 2,786 Tigrinya, 187 Tigre and 2 Ge’ez evaluation words. For Tigre and Ge’ez the resource is the Tigrinya one, since the Tigrinya MoVoC-Tok scores them cross-lingually.
8. Table 2: held-out evaluation-set sizes and reconstructed MorphScores for all three tokenizers; the comparison with the fusional-language mean is withdrawn.
9. Table 3: withdrawn, together with the OPUS appendix; this version reports no translation results.
10. Table 4: held-out precision and entropy for all three tokenizers; “best in all four languages” corrected to two of four on precision. A new table reports bootstrap confidence intervals for the differences against BPE.
11. Section 7: segmentation examples replaced by actual output; translation example withdrawn.
12. Sections 8 and Limitations: conclusions aligned with the corrected results.
13. Table 5: realised vocabulary sizes (114,553, not 152k).
14. Table 6: examples replaced by records present in the released data.